\pdfoutput=1

\documentclass[11pt]{article}
\usepackage{graphicx}

\usepackage{naacl2021}

\usepackage{times}
\usepackage{latexsym}

\usepackage[T1]{fontenc}

\usepackage[utf8]{inputenc}

\usepackage{microtype}

\title{Transfer learning from High-Resource to Low-Resource Language Improves Speech Affect Recognition Classification Accuracy}

\author{Sara Durrani \\
  Fast Nuces\\
  Islamabad, Pakistan.\\
  \texttt{sara.durrani@nu.edu.pk} \\\And
  Umair Arshad \\
  Fast Nuces\\
  Islamabad, Pakistan.\\
  \texttt{umair.arshad@nu.edu.pk} \\}

\begin{document}
\maketitle
\begin{abstract}
Speech Affect Recognition is a problem of extracting emotional affects from audio data.  Low resource languages corpora are rear and affect recognition is a difficult task in cross-corpus settings. We present an approach in which the model is trained on high resource language and fine-tune to recognize affects in low resource language. We train the model in same corpus setting on SAVEE, EMOVO, Urdu, and IEMOCAP by achieving baseline accuracy of 60.45, 68.05, 80.34, and 56.58 percent respectively. For capturing the diversity of affects in languages cross-corpus evaluations are discussed in detail. We find that accuracy improves by adding the domain target data into the training data.  Finally, we show that performance is improved for low resource language speech affect recognition by achieving the UAR OF 69.32 and 68.2 for Urdu and Italian speech affects.
\end{abstract}
\textbf{Index Terms}: Transfer learning, cross-corpus, Affect Recognition
\section{Introduction}

Speech affect analysis is an open research problem that is making an impact on the research community. For many years, work is being done for speech recognition and detection (\citealp{karsten2007axiomatic}). The automatic speech recognition systems work by identifying different speakers (\citealp{javed2020alphalogger}). The models developed can be speaker-dependent or independent. These acoustic models have good results and fulfill several domain requirements of different applications (\citealp{tariq2019accurate}). The problem arises when these models have to deal with speech data that comes from speakers of different age groups, gender, accent, and language (\citet{asad2020deepdetect}). In this context, language is a huge barrier for a lot of speech recognition systems. In real-time scenarios, speech includes different affects that make a huge impact on the performance of speech recognition systems (\citet{dilawar2018understanding}). These affects include sadness, anger, disgust, happiness, and many more (\citet{beg2010graph}). The speech affect has different applications in multiple domains that include call centers (\citealp{burkhardt2006detecting}), face affect recognition (\citealp{jain2018hybrid}), smart classrooms, human behaviour analysis, and increasing customer shopping experience (\citealp{vidrascu2005annotation}). The affects in speech are also being used to track depression and mental pressure (\citealp{bangash2017methodology}) in different smart home and offices environments (\citealp{huang2019speech}).\\
In the early work, we have found the continuous Hidden Markov Model and Gaussian Mixture Model. The paper (\citealp{schuller2003hidden}) discusses two methods, first is about a global static framework that extracted derived features from the speech signal (\citealp{javed2020collaborative}). The second method extracts low-level instantaneous features by applying the continuous Hidden Markov Model.

Feature engineering has become more advanced for capturing a lot of information that became part of different machine learning techniques used for speech affect recognition or analysis (\citealp{uzair2019weec}). Support vector machine (\citealp{pan2012speech}) is trained using these useful features that include linear predictive spectrum (LPCC), Mel-frequency spectrum coefficients (MFCCs) (\citealp{logan2000mel}), speech energy, and pitch by achieving 91 percent accuracy on the Chinese database (\citealp{sahar2019towards}). The speech affect recognition accuracy has been boosted with the advent of deep neural architectures (\citealp{beg2013constraint}). The automatic feature selection property of networks made the achievement of this task easy (\citealp{awan2021top}). The deep neural architecture (\citealp{han2014speech}) extract high-level features and produce probability distributions for deep neural networks. The extreme learning machine which is a single layer hidden network feeds utterance level features and identify the hidden emotions (\citealp{naeem2020deep}). Further to this, deep neural networks are explored along with LSTMs for this speech task (\citealp{zhao2019speech}). The 1d and 2d convolution neural network has learned global and local affect features from speech and spectrograms (\citet{alvi2017ensights}). This architecture consists of two learning feature blocks: a max-pooling layer, and one convolutional layer. Overall the architecture of the network takes advantage of both LSTM and CNN (\citealp{zafar2020search}). The LSTM networks are explored for this task in depth for capturing more enhancing features and improving the accuracy. In (\citealp{xie2019attention}), the attention mechanism is introduced with LSTM for processing the time-series signal information. This work increased the accuracy of the standard emotion corpus (\citealp{khawaja2018domain}). \\

Attention based Bidirectional LSTM with multi stream is proposed in (\citealp{chiba2020multi}) for individual temporal speech parameters. An advanced form of convolutional network named temporal convolutional network is introduced in (\citealp{liu2020temporal}) with a vector quantization variational encoder. The encoder is trained in unsupervised manner with a lot of unlabeled data. Due to the evaluations on a single corpus, the concept of transfer learning is proposed and used widely (\citealp{arshad2019corpus}). The Siamese neural network's loss is modified (\citealp{DBLP:conf/aaai/FengC20})for training in transfer learning setting. The results are achieved by the distance loss between same and different classes. In addition to this, problems arise when the systems are tested for cross language data. This work (\citealp{liu2020cross}) handles the problem of cross corpus speech affect recognition. The architecture consists of two modules: first model features are extracted and domain adaptive layer is introduced. The cross corpus experimentation and evaluations effect the performance of affect recognition systems (\citet{farooq2019melta}). \\

Now, a variety of speech corpora exist that have data in different languages, advanced emotions, and diverse labeling. A variety of populations speak different languages. On record, *389 languages \footnote{\url{https://www.mustgo.com/worldlanguages/world-languages/}} are spoken by one million people in different areas of the world that make 94.1 percent of the world's population.  This seems very difficult to have large data sets of every language for model training (\citealp{zafar2019using}). The data seems to be inadequate for ever low-resource language. The researchers and speech system developers face the problem of having not enough data (\citealp{javed2019fairness}). Therefore, it seems impossible to have a single model trained on a single language corpus to capture all variations, dynamics, recognition, emotions, and affects (\citealp{beg2019algorithmic}). The generalization of the model is required to make it work effectively.\\

\begin{figure}[h]
\includegraphics[width=7.5cm, height=3cm]{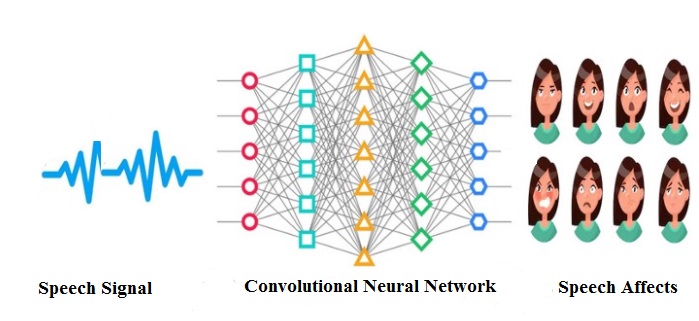}
\caption{The picture shows a general view of speech affect recognition as speech signal inputs into a CNN and recognizes the different speech affects.}
\end{figure}
In automatic speech recognition, it is mostly not taken into account that if trained on a single corpus the model will fail in cross-corpus settings (\citealp{naeem2020subspace}). For the resolution of this problem, transfer learning is introduced into the deep neural architectures. Transfer learning transfers the domain knowledge from the source to the target domain. This technique seems very helpful where very less labeled or non-labeled data is present (\citealp{majeed2020emotion}). The accuracy rate increases due to transfer learning in speech recognition for low-resource languages. Due to speech features capturing capacity and adaptability to perform in a cross-corpus setting, we have used a encode-decoder model with attention (\citealp{zahid2020roman}). Transfer learning is applied on convolutional neural networks followed by LSTM with an attention mechanism. In this work, we have solved the above-mentioned challenges by analyzing the results in same corpus, cross-corpus and multilingual settings.

\section{Methodology}

We have used the encoder-decoder model with attention (\citealp{bansal2018low}) for all experiment settings to transfer the domain knowledge. This has allowed us to transfer the training parameters between chosen models. The form of learning introduced through transfer learning is really flexible and transfers all domain knowledge from high resource to a low resource one. The hyper parameters can also set to make it easy to fit into the available computation resources. We train a English model on IEMOCAP data set to make it available for different evaluations and transfer of parameters. Further for Affect SAR Model (IEMOCAP-Urdu), we have used IEMOCAP pre-trained model and retrained it with 320 samples of Urdu. During training the parameters are updated that helps in transferring domain knowledge from high resource language to low resource one. Only encoder parameters are updated due to the reason that speech signal knowledge is same and can be processed in a same way.

\begin{figure}[h]
\includegraphics[width=7.5cm, height=7cm]{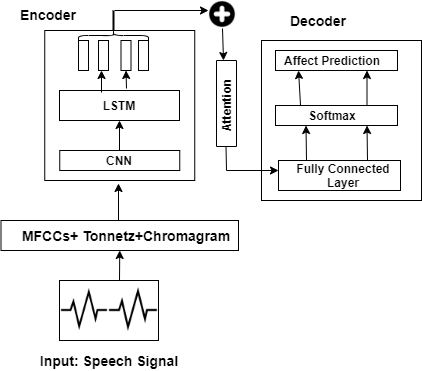}
\caption{The encoder-decoder architecture for both speech affect recognition. The speech signal is the input of encoder and decoder generates class labels. }
\end{figure}

\section{Experimental Setup}
\subsection{Speech Data sets}
For the task of speech affect recognition in different languages, we have selected four different publicly available speech data sets of different languages for capturing the maximum diversity. These data sets are annotated differently and have recordings for basic and advanced speech affects. The affects are studied in depth by considering the important positive and negative classes for this classification problem. The table 1 shows the details for different selected data sets.

\begin{table*}
\centering
\begin{tabular}{ |p{1.6cm}|p{1.8cm}|p{1.8cm}|p{6cm}|p{4cm}|  }
\hline
\textbf{Data Set} & \textbf{Language} & \textbf{Recordings} & \textbf{Affects List} & \textbf{References} \\
\hline
SAVEE & English & 480 & Anger, Sad, Neutral, Happy, Surprise,fear, Disgust  & (\citealp{jackson2014surrey})\\
EMOVO & Italian & 588 & Anger, Sad, Neutral, Joy, Surprise,fear, Disgust  & (\citealp{costantini2014emovo}) \\
URDU & Urdu & 400& Anger, Sad, Neutral, Happy  & (\citealp{latif2018cross})\\
IEMOCAP & English & 5531 & Anger, Sad, Neutral, Happy, Excited  & (\citealp{busso2008iemocap}) \\
\hline
\end{tabular}
\caption{\label{citation-guide}
The table shows corpora names, language, no of utterances, the recorded affects and references. The class labels are assigned according to the genre of affects.
}
\end{table*}

All datasets are relevent and prepared in a well-organized way. IEMOCAP has most of the data and it consists of five sessions in the form of audio, video and images. The sessions are recorded by five pairs of males and females. The gold labels are assigned to data by crowd sourcing. The data is gathered in controlled environment as acted under specific conditions, thus obtain high accuracy results.
\subsection{Speech Features Details}
The extraction of features is a very crucial part for developing a good model. We use Chromagram and Tonnetz representation feature sets that are highly used for the distinguishable representation of pitch and harmony. The feature specially spectral contrast shows detailed spectra of sound in contrast to MFCCs spectograms. Tonnetz captures pitch and harmony classes of sound. The tonal centroids of sound are measured in tonal centroid space (\citealp{harte2006detecting}).The pitch classes along harmonic relations are studied in depth in the form of Harmonic network representations. This features set includes frequency, spectral information, pitch, energy, static and dynamic variations.
\section{Experimental Results}
This section explains the different cross-corpus settings and comparisons of our results with the previous results of existing techniques. The different possible scenerios are explored and studied in detail.

\subsection{Proposed Baseline Model Results for single corpus}

We trained our model using proposed approach on each corpus that set the baseline accuracy results. The performance of our approach is compared with the very relevant work using Deep Belief Networks(\citealp{latif2018transfer}) and sparse autoenoder with SVM using transfer learning in speech Emotion Recognition (\citealp{deng2013sparse}). This enables us to train model on 80 percent and test it on 20 percent unseen data. The figure 3 shows the comparison results and it is evident that our approach has outperformed the other two with accuracy improvement.      

\begin{figure}[h]
\includegraphics[width=7.5cm, height=4.5cm]{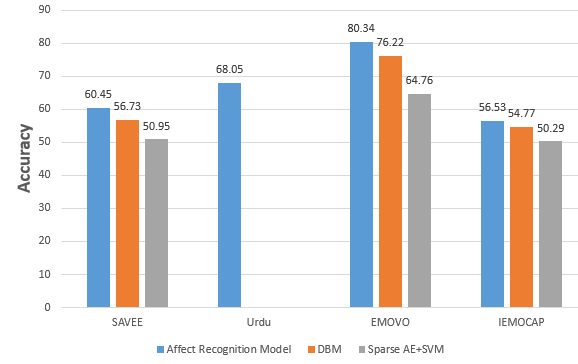}
\caption{The baseline accuracy comparison of different corpora on our Affect Recognition Model, DBM and Sparse auto-encoder with SVM.}
\end{figure}

\subsection{Cross Corpus Setting}
In a cross-corpus setting, we have used IEMOCAP and EMOVO for training the model. The overall task is to use one language data set for training and the remaining corpora for testing. The remaining corpora which include SAVEE and Urdu are used for evaluations. The cross-corpus settings are fairly good for generalizing the language barriers as this evaluation makes models strong and provides space for improvement of results. We have compared the recognition rate of different languages and obtained the results. Figure 4 shows the recognition rate for our model and the other two existing approaches.

\begin{figure}[h]
\includegraphics[width=7.5cm, height=4.5cm]{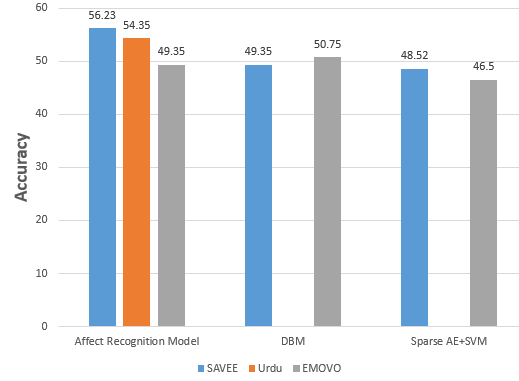}
\caption{The comparison results of approaches in a cross-corpus setting when the Affect Recognition Model is trained using the IEMOCAP data set.}
\end{figure}

\begin{figure}[h]
\includegraphics[width=7.5cm, height=4.5cm]{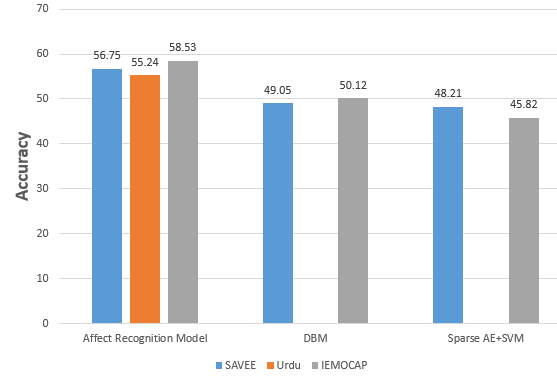}
\caption{The comparison results of approaches in a cross-corpus setting when the Affect Recognition Model is trained using the EMOVO data set.}
\end{figure}

It is evident by the obtained results that the Affect Recognition Model outperforms the existing two by the improvement in accuracy and performs better in cross-corpus settings.

\subsection{Transfer Learning Results}

The concept of transfer learning enables us to use corpora of different languages jointly for training. This makes the performance of models better for different corpora. We use IEMOCAP and Urdu for training and tested the model for EMOVO,SAVEE and Urdu. The Urdu data set is also divided 80 percent for testing and 20 percent for training. IEMOCAP has large set of data, we utilized three sessions as training and rest left for evaluations. The evaluations are done using three fold cross validation for testing specified corpora i.e. EMOVO,SAVEE and Urdu.
\subsection{Evaluations}
\textbf{UAR: }The unweighted average recall score is also calculated for Speech Affect Recognition. Unweighted Average Recall is a parameter that is calculated for recall of every class. It gives out an easy calculation for data set accuracy when the data set samples count is imbalanced as compared to all other classes. The table 2
shows the results of transfer learning which depicts that for Urdu, English and Italian it has achieved 69.32, 68.52 and 68.52 UAR. Therefore, it can be seen that results can be obtained for the languages that have less annotated data set and least training capacity. This transfer learning approach plays its part in providing more assistance
in less resources.

\begin{figure}[h]
\includegraphics[width=7.5cm, height=5cm]{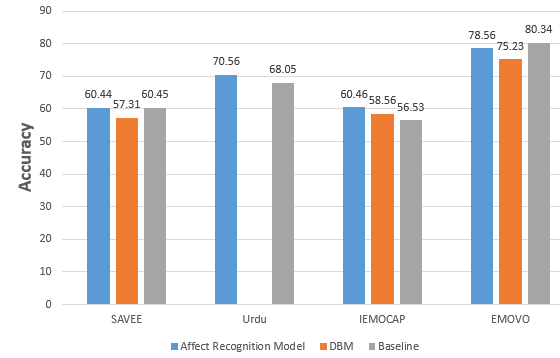}
\caption{The comparison results of approaches in a transfer learning setting when the affect recognition model(IEMOCAP-Urdu) is trained.}
\end{figure}
\begin{table}
\centering
\begin{tabular}{lc}
\hline
\textbf{Data Set} & \textbf{UAR}\\
\hline
SAVEE & 68.52 \\
EMOVO &  70.41\\
Urdu &  69.32\\ 
\hline
\end{tabular}
\caption{The UAR score of selected corpora}
\label{tab:accents}
\end{table}

\section{Analysis}
From different experiments that we have performed for the Speech Affect Recognition has allowed us to note key points. The transfer learning outperformed the baseline results within same corpus training. The accuracy results for all data sets are higher in transfer learning settings as compared to cross-corpus and baseline( even when the model is trained and tested on the same corpus). Low resource languages in our study, Urdu, and Italian have scored high classification accuracy rates while comparing with high resource language English. The model encoder parameters along with attention parameters are all transferred that proved to be the most effective.The transfer of alone decoder parameters do not improve the efficiency for increase in accuracy. The dependency has created and it sums that decoder parameters without encoder might not transfer enough knowledge. \\
We have also found that adding the domain data while training improves the classification accuracy rate. However, the performance of the systems and models drops due to different associated factors. Speech Affects are highly sensitive to age, gender, noise, and language diversity. We have studied that transfer learning solves this problem to a great extent but for more accuracy, the domain data can be added to the training data from a variety of languages.

\section{Conclusions and Future Work}
In this paper, we evaluate the performance of encoder-decoder model with attention for Speech Affect Recognition in the same corpus, cross-corpus, and transfer learning settings. The detailed experiments show that the transfer learning based model outperformed the existing approaches for mentioned settings. We perform on four different language corpora by transferring high resource language features and domain knowledge to low resource languages. This would be very helpful in building applications specific to Speech Affect Recognition. Moreover, this technique also solves a problem when less or non-labeled low resource language data is available. We show that it is possible to improve the results by this approach. In our future work, we aim to work for advanced speech affects in low resource languages to capture the high diversity and improving the Speech Affect Recognition rate.

\bibliography{naacl2021,custom, References}
\bibliographystyle{acl_natbib}

\end{document}